\documentclass[letterpaper, 10 pt, conference]{ieeeconf}  %

\IEEEoverridecommandlockouts                              %
\usepackage[utf8]{inputenc}
\usepackage[T1]{fontenc}
\usepackage{mathptmx} %
\usepackage{times}    %
\usepackage{cite}
\usepackage{amsmath,amssymb,amsfonts}
\usepackage{algorithmic}
\usepackage{graphicx}
\usepackage{textcomp}
\usepackage[table]{xcolor}
\usepackage{booktabs}
\usepackage{multirow}
\usepackage[font=small]{caption}
\usepackage[hidelinks]{hyperref}
\usepackage[capitalise,noabbrev]{cleveref}
\crefname{table}{Table}{Tables}
\Crefname{table}{Table}{Tables}
\crefname{figure}{Fig.}{Figs.}
\Crefname{figure}{Fig.}{Figs.}
\crefname{equation}{Eq.}{Eqs.}
\Crefname{equation}{Eq.}{Eqs.}
\def\BibTeX{{\rm B\kern-.05em{\sc i\kern-.025em b}\kern-.08em
    T\kern-.1667em\lower.7ex\hbox{E}\kern-.125emX}}

\newcommand{\msection}[1]{\section{#1}\vspace{-1.0mm}}
\newcommand{\OurMethod}{\textsc{ForceTwin}}

\newcommand{\sd}[2]{#1\,{\tiny$\pm$#2}}
\newcommand{\tabtitle}[1]{{\scshape\bfseries #1}\upshape}

\newcommand{\PAR}[1]{\vskip 0pt \noindent{\bf #1~}}

\newif\ifprintbibliography
\printbibliographytrue

\begin{document}

\title{\LARGE \bf ForceTwin: Physics-informed Digital Twins for Robotic Manipulation from Instrumented Human Interaction}

\author{
Tim Engelbracht$^{1,\dagger}$\quad René Zurbrügg$^{1}$\quad Mayank Mittal$^{1,2}$\quad Marco Hutter$^{1}$ \\
Marc Pollefeys$^{1,3}$\quad Hermann Blum$^{4}$\quad Zuria Bauer$^{1}$ \\[0.3em]
{\normalsize $^{1}$ETH Zurich\quad $^{2}$NVIDIA\quad $^{3}$Microsoft\quad $^{4}$University of Bonn}
\thanks{$^{\dagger}$Corresponding author: {\tt\small tengelbracht@ethz.ch}}
}

\makeatletter
\def\@IEEEaftertitletext{%
  \begin{center}
    \includegraphics[width=0.86\textwidth]{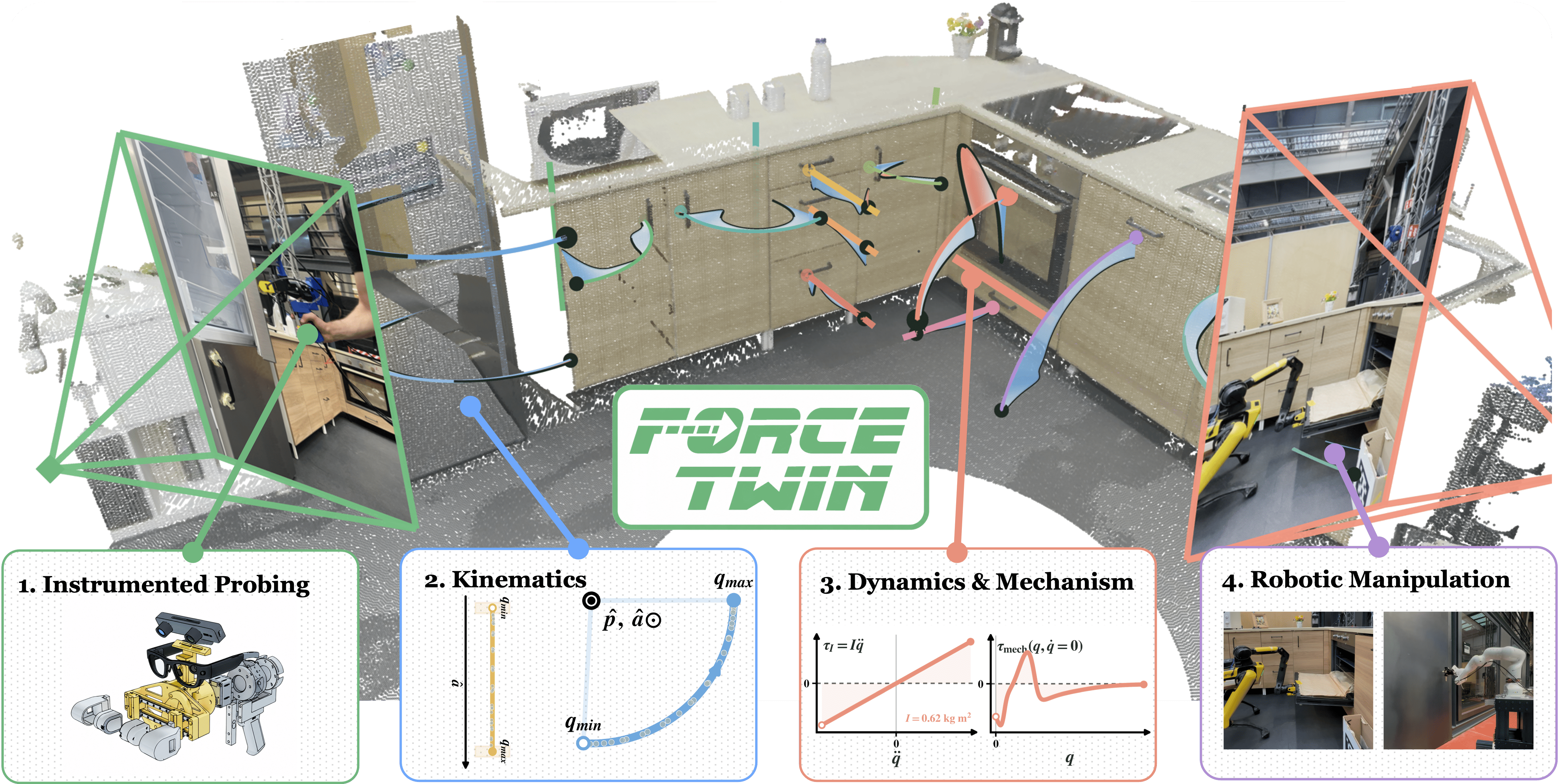}
    \refstepcounter{figure}\label{fig:teaser}%
    \parbox{1.0\textwidth}{\footnotesize\vspace{4pt}
      \textbf{Fig.~\thefigure.} \textbf{\OurMethod{} builds a physics-informed digital twin from instrumented human interaction.} Interaction trajectories estimate object kinematics, while measured wrenches identify dynamics and mechanism behavior. Combined with semantic 3D scene reconstruction, \OurMethod{} supports dynamics-aware robot control and high-fidelity simulation parameters for reinforcement learning.
}
  \end{center}
}
\makeatother

\maketitle
\begin{abstract}
Manipulating objects requires understanding not only their motion, but also the physical properties that determine it. For articulated objects, these include inertia, friction, and mechanisms such as springs or door closers, whose effects can vary with configuration and velocity. Such properties are not directly observable from appearance: visually identical doors may require very different effort to manipulate. Existing digital-twin pipelines recover primarily kinematics or assign static physical parameters from visual and language priors, which can yield physically implausible estimates. As a result, state-dependent mechanism dynamics remain unidentified and are not represented in standard asset formats. We present \OurMethod{}, a system for identifying physics-informed digital twins of articulated objects from instrumented human interaction. A person probes an object using a handheld force-sensing gripper, providing synchronized poses and interaction forces from which we estimate the articulation, parametric dynamics including inertia, Coulomb friction, viscous damping, and a structured neural residual capturing state-dependent mechanism forces. \OurMethod{} nearly halves the inertial-parameter error of a VLM prior. As a feedforward dynamics model for impedance control on a Spot and a Franka FR3, \OurMethod{} achieves 87\% goal completion across nine object-embodiment pairs, compared with 60\% using VLM-prior and 57\% using kinematics-only twins, with the largest gains on objects whose strong mechanisms cause both baselines to stall. We further use the identified twins to train whole-body door-traversal policies and deploy them in the real world. Project page: \href{https://timengelbracht.github.io/forcetwin-website/}{timengelbracht.github.io/forcetwin-website}
\end{abstract}

\section{Introduction}
Robotic manipulation requires more than knowing how an object can move. To open a drawer, a dishwasher, or a door with a closer, a robot must also account for the forces induced by inertia, friction, and internal mechanisms. These dynamics can vary substantially between visually similar objects and are often impossible to infer from appearance alone: the same drawer may be empty or heavily loaded, while a concealed closer can make an otherwise ordinary door strongly resist motion. A digital twin for manipulation must therefore capture not only an object's geometry and articulation, but also the instance-specific dynamics that govern its interaction.

These dynamics have to be measured. Appearance provides at best a prior, while motion alone identifies inertia, friction, and mechanism torque only up to a common scale that the contact wrench resolves. Feedback cannot simply replace the missing model: strong resistance may indicate that a drawer is stuck and the robot should stop, or be the expected behavior of a spring-loaded door that requires additional effort. %

Existing twin pipelines neither measure this effort nor model its state-dependent component. Reconstruction methods recover kinematic structure%
~\cite{gu2026artisgfunctional3dscene,büchner2026articulated3dscenegraphs,Huang_2026,yu2026pandoraarticulated3dscene}, while recent methods assign physical parameters from visual and language priors~\cite{iliash2026artiverse,li2026physgraphphysicsaware3dscene}.
Robot-probing approaches~\cite{katz2008interactive,martinmartin2022coupled,buchanan2026online} measure interaction, but recover at most constant friction models and require robot deployment. None identify the nonlinear, state-dependent response of mechanisms such as door closers or spring-loaded catches, which no set of constant parameters can describe, and standard asset formats provide no representation for this response.

\begin{figure*}[!t]
  \centering
\includegraphics[width=0.9\linewidth]{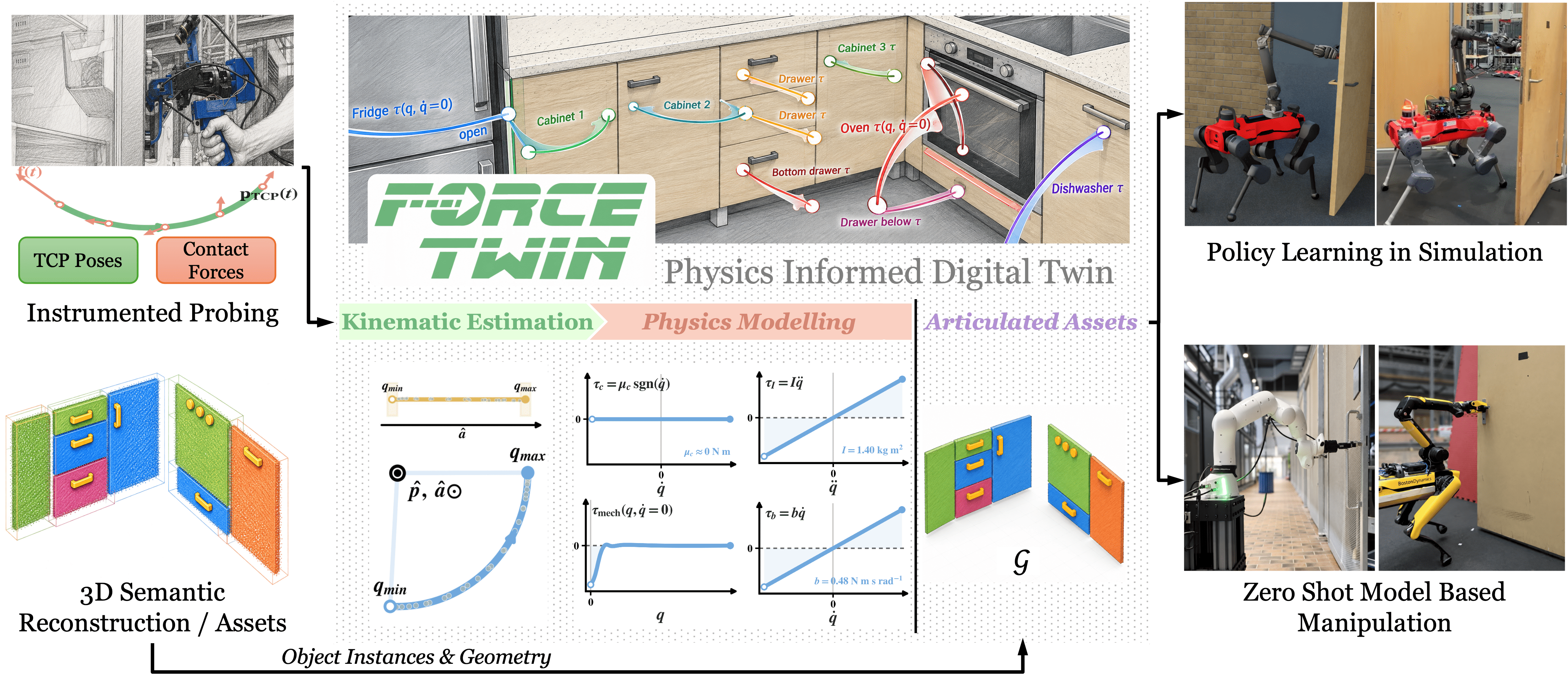}
  \vspace{-5pt}
  \caption{\textbf{The \OurMethod{} pipeline.} Instrumented human interaction provides
    synchronized tool poses and contact wrenches from which we identify an
    articulation model and its generalized-effort dynamics. Geometry is
    reconstructed from the same recording or supplied independently, and the
    physical model is registered to it to produce a twin for robot control
    or simulation.}
    \vspace{-15pt}
  \label{fig:pipeline}
\end{figure*}

We take a different route (\cref{fig:pipeline}): a person measures the object. Handheld force-sensing grippers have made in-the-wild interaction capture practical~\cite{Engelbracht_2026_CVPR,choi2026inthewildcompliantmanipulationumift}; we repurpose them for system identification. With \OurMethod{}, a person probes an articulated object, and from the tool trajectory and contact wrenches we identify the object's kinematics and dynamics (\cref{fig:teaser}), decoupling identification from robot deployment. Humans can reach objects without a robot present, safely excite their full range across varied speeds, and do so faster than contact-based robot exploration.
We model joint effort semi-parametrically. Inertia, Coulomb friction, and viscous damping are estimated under nonnegativity constraints, yielding physically valid parameters that map directly to simulators and controllers. A structured neural residual captures the remaining state-dependent mechanism response. The resulting twin can be used directly for model-based control or exported as a simulation asset. \\

In summary, our contributions are:
\begin{itemize}
  \item We introduce \OurMethod{}, which identifies instance-specific articulated-object dynamics from instrumented human interaction, including nonlinear, state-dependent mechanism responses.%
  \item We formulate identification as a semi-parametric partially linear model, combining physically constrained inertia and friction parameters with a %
  structured neural residual, yielding simulation-compatible parameters and a predictive model of joint effort.
  \item We show that one identified twin serves both control and simulation. As the feedforward model of an impedance controller, it reaches 87\% goal completion across nine object--embodiment pairs, where prior-based and kinematics-only twins reach 60\% and 57\% and stall entirely on objects with strong mechanisms; incorporated into simulation assets, the identified dynamics support policy learning, where we train whole-body door-traversal policies and deploy them in the real world.
\end{itemize}

\msection{Related Work}

\PAR{Articulated Object Perception and Kinematics.} Estimating articulation models by acting on the world is a classic instance of interactive perception~\cite{Bohg_2017}: a robot pushes an object and infers its kinematic model from the observed motion~\cite{katz2008interactive,Sturm_2011}. Later work removes the interaction, estimating articulation from a single RGB-D frame~\cite{sun2023opdmultiopenabledetectionmultiple}, from two observed object states~\cite{liu2023paris}, with learned category-independent models~\cite{jain2021screwnet}, or from in-the-wild video~\cite{werby2025articulatedobjectestimationwild}, and scene-graph pipelines lift such estimates to scene scale for robotic manipulation~\cite{büchner2026articulated3dscenegraphs,yu2026pandoraarticulated3dscene,gu2026artisgfunctional3dscene,fu2026funfactbuildingprobabilisticfunctional}. All of these methods estimate kinematic structure, i.e., joint types and axes, but not the dynamic parameters that govern interaction forces; our kinematics stage builds on this line of work, adopting the twist formulation of works such as~\cite{jain2021screwnet,buchanan2023onlineestimationarticulatedobjects} and interactive BIC-based model selection of~\cite{Sturm_2011}, but treats the kinematic model as the input to dynamic identification rather than the final result.

\PAR{Instrumented Interaction and Physical Model Identification.} Closest to our setting are methods that identify physical parameters from forceful interaction. Door inertia and a velocity-dependent deceleration profile have been identified from robot interaction and released-door trajectories~\cite{endres2013dynamics4door}, and human opening forces on doors and drawers have been captured with an instrumented hook, yielding force-versus-angle profiles used for haptic recognition and monitoring~\cite{jain2013forces}. We share their premise that dynamics must be measured, but differ in what is recovered and for what: our semiparametric formulation yields interpretable, simulation-ready parameters plus a nonlinear mechanism model, rather than a per-object regression, and our capture requires only a handheld gripper rather than a robot. Robot-mounted estimators also go beyond kinematics: coupled recursive filters over vision, wrist forces, and proprioception additionally estimate a quasi-static Coulomb friction model, i.e., stiction and constant kinetic friction, while deliberately neglecting inertial and viscous effects~\cite{martinmartin2022coupled}, and factor-graph estimators use force factors to disambiguate the articulation but identify no dynamic parameters~\cite{buchanan2026online}. Our identification retains all of these terms, inertia, Coulomb and viscous damping, under physical constraints, and adds the nonlinear mechanism response. Ditto~\cite{jiang2022ditto} builds ``digital twins of articulated objects from interaction,'' but its twins carry default physical parameters, while the dynamics parameters are not identified. Finally, handheld instrumented grippers~\cite{chi2024universalmanipulationinterfaceinthewild,Engelbracht_2026_CVPR,choi2026inthewildcompliantmanipulationumift} have made in-the-wild interaction capture practical, so far primarily to record demonstrations for policy learning; we repurpose this hardware for system identification.

\PAR{Real-to-Sim and Simulation Asset Creation.}
A complementary line of work turns real scenes into simulation-ready articulated assets: from images alone~\cite{chen2024urdformer}, from a robot push~\cite{ma2023sim2real2}, through human-in-the-loop annotation~\cite{torne2024rialto}, or by retrieving similar ``digital cousins'' from an asset database~\cite{dai2024automatedcreationdigitalcousins}. In all four, the resulting physics is defaulted, annotated, or inherited from the retrieved asset. A parallel thread assigns physical parameters from visual and language priors~\cite{iliash2026artiverse,li2026physgraphphysicsaware3dscene,Huang_2026}, fits them to visual tracking~\cite{jiang2025phystwin}, or tunes simulator distributions against real task rollouts~\cite{chebotar2019simopt}. None of these approaches measure the interaction forces of the specific instance being modeled. \OurMethod{} complements this line of work with per-instance dynamics, identified from force measurements and exported in a format such pipelines can consume.

\msection{Method}

\subsection{Problem Formulation}
We consider a rigid articulated part with a single degree of freedom and
identify its kinematics and interaction dynamics from instrumented human
probing. The observations consist of two synchronized streams,
\begin{equation}
  \mathcal{D}
  =
  \left\{
    {}^{W}T_{\mathrm{TCP},i},
    {}^{W}\mathcal{F}_{i}
  \right\}_{i=1}^{N},
\end{equation}
where ${}^{W}T_{\mathrm{TCP},i}$ is the pose of the tool center point (TCP)
in a world frame $W$, and ${}^{W}\mathcal{F}_{i}\in\mathbb{R}^{6}$ is the contact wrench
applied by the tool to the object after compensating for the weight and
inertial loading of the instrumented gripper. We assume that the tool
maintains a rigid, non-slip grasp and that the articulated part follows an
ideal revolute or prismatic joint during the retained measurements.
Unless marked otherwise, all quantities are expressed in $W$, and bold
denotes a stacked multi-component quantity.
From $\mathcal{D}$, we estimate the articulation class
$\kappa\in\{\mathrm{revolute},\mathrm{prismatic}\}$, the joint twist
${}^{W}\boldsymbol{\xi}$, the observed range
$\mathcal{Q}=[q_{\min},q_{\max}]$, and a generalized-effort model $  \hat\tau
  =
  \hat\tau
  (q,\dot q,\ddot q).$
We represent the latter by simulator-compatible effective parametric coefficients
$\boldsymbol{\beta}=[\mu,b,I]^\top$ and a structured mechanism residual
$\mathcal{M}_{\mathrm{mech}}=\{g_\theta,c_\phi\}$. Together, these quantities
form the physical model
\begin{equation}
  \mathcal{M}_{\mathrm{phys}}
  =
  \left(
    \kappa,\,
    {}^{W}\boldsymbol{\xi},\,
    \mathcal{Q},\,
    \boldsymbol{\beta},\,
    \mathcal{M}_{\mathrm{mech}}
  \right).
\end{equation}
The model is estimated independently of a particular geometric reconstruction.
A physics-informed digital twin is obtained  afterwards by registering and
associating $\mathcal{M}_{\mathrm{phys}}$ with a geometric asset or a robot's
object representation.

\subsection{Kinematics Estimation}
Under the rigid-grasp assumption, the tracked TCP positions
$\{p_i\}_{i=1}^{N}$ follow the one-dimensional motion of the articulated
part. We represent this motion using a screw in product-of-exponentials
form~\cite{jain2021screwnet,Sturm_2011,lynch2017modern},
\begin{equation}
  \label{eq:screw}
  \bar p_i
  =
  \exp\!\left(
    [{}^{W}\boldsymbol{\xi}]^\wedge q_i
  \right)
  \bar p_0 ,
\end{equation}
where $\bar p_i\in\mathbb{R}^{4}$ denotes the homogeneous TCP position,
$q_i$ is the scalar joint coordinate, and
${}^{W}\boldsymbol{\xi}=[\omega^\top,v^\top]^\top\in\mathbb{R}^{6}$ is the
joint twist, and $[\cdot]^{\wedge}$ maps a twist to its $4\times4$ matrix
form in $\mathfrak{se}(3)$. We remove the scale ambiguity between
${}^{W}\boldsymbol{\xi}$ and $q$ by imposing $\|\omega\|_2=1$ for revolute
joints and $\|v\|_2=1$ for prismatic joints. We set the coordinate origin at
the first retained configuration and orient the axis so that the first
sustained displacement has positive $q$. These conventions fix the units and
signs of the subsequently identified dynamic parameters.
A revolute candidate satisfies $\omega\neq0$ and
$\omega^\top v=0$, while a prismatic candidate satisfies $\omega=0$.
We fit both candidates and select the articulation class using the Bayesian
information criterion~\cite{Schwarz1978,Sturm_2011}. We then refine the
selected model in a factor graph, similar
to~\cite{buchanan2026online}, jointly estimating the joint geometry and the
per-sample states $\{q_i\}$. For revolute joints, the refinement enforces
zero screw pitch. The observed extrema of the refined states define
$\mathcal{Q}=[q_{\min},q_{\max}]$. %

\subsection{Dynamics Identification}
\PAR{Wrench Compensation and Generalized Effort.}
The force-torque sensor measures both the contact wrench and the wrench
generated by the instrumented gripper itself. We transform the measured
wrench into the world frame and subtract the calibrated gravitational and
inertial wrench of the gripper, yielding the estimated contact wrench
${}^{W}\mathcal{F}_i$ applied by the tool to the object. %
We order spatial twists as angular followed by linear velocity and wrenches
as moment followed by force. The kinematic estimate then defines the object
Jacobian at the TCP,
\begin{equation}
  \label{eq:objjac}
  J_{\mathrm{obj}}(q_i)=
  \begin{cases}
    \begin{bmatrix}
      \omega\times(p_i-o),\, \omega
    \end{bmatrix}^{T},
    & \kappa=\mathrm{revolute},
    \\[2pt]
    \begin{bmatrix}
      v,\, \mathbf{0}
    \end{bmatrix}^{T},
    & \kappa=\mathrm{prismatic}.
  \end{cases}
\end{equation}
where $o$ is a point on the revolute axis, and $J_{p}$ denotes the linear
block of $J_{\mathrm{obj}}$, the three rows that map $\dot q$ to the TCP
velocity. By virtual work, the generalized
effort applied by the tool is $  \tau_i
  =
  J_{\mathrm{obj}}(q_i)^\top
  {}^{W}\mathcal{F}_i .$
Under the rigid-grasp and ideal-joint assumptions, wrench components that
perform no virtual work along the permitted articulation lie in the null
space of $J_{\mathrm{obj}}^\top$ and do not contribute to $\tau_i$. The
projection does not remove errors caused by wrench calibration, joint-axis
estimation, grasp compliance, or slip; these effects remain part of the
measurement residual.
We obtain $\dot q_i$ and $\ddot q_i$ by differentiating a smoothed joint-state
trajectory. Samples near joint limits, with negligible velocity, or with
implausibly large differentiated accelerations are removed using excitation gates.

\PAR{Generalized-Effort Model.}
We define $\tau_i$ as the generalized effort applied to the object and model
the effort required to realize the observed motion as
\begin{equation}
  \label{eq:dyn-model}
  \tau
  =
  \underbrace{
    I\ddot q
    + \mu\,\operatorname{sgn}(\dot q)
    + b\dot q
  }_{\text{effective parametric dynamics}}
  +
  \underbrace{
    g_\theta(q)
    + c_\phi(q,|\dot q|)\dot q
  }_{\text{structured mechanism residual}}
  +
  \varepsilon ,
\end{equation}
where $I$ denotes rotational inertia for revolute joints and effective mass
for prismatic joints, $\mu$ is the Coulomb friction magnitude,
$b$ is viscous damping, and $\varepsilon$ contains measurement noise and
unmodeled effects. We write
$\tau_{\mathrm{mech}}(q,\dot q)=g_\theta(q)+c_\phi(q,|\dot q|)\dot q$ for the
structured residual as a whole.

The sign convention in \cref{eq:dyn-model} describes the effort that must be
applied to move the object. Consequently, the compensating damping effort is
$+c_\phi\dot q$. The corresponding damping effort exerted by the object is
$-c_\phi\dot q$, whose mechanical power is
\begin{equation}
  P_{\mathrm{damp}}
  =
  -c_\phi(q,|\dot q|)\dot q^2
  \leq 0
\end{equation}
when $c_\phi\geq0$.
The function $g_\theta(q)$ captures configuration-dependent generalized
loads. For joints whose motion is orthogonal to gravity, this term primarily
represents mechanisms such as door closers and spring-loaded elements. For a
horizontal revolute axis, it may also contain gravitational torque unless
gravity is compensated using an independently available mass and
center-of-mass estimate. We therefore interpret $g_\theta$ as a
configuration-dependent load rather than claiming that it uniquely isolates
a mechanical spring.
\PAR{Residual Fitting.}
We first estimate the effective parametric coefficients using samples with sufficient motion and force excitation. For sample $i$, define
\begin{equation}
  z_i
  =
  \begin{bmatrix}
    \operatorname{sgn}(\dot q_i) &
    \dot q_i &
    \ddot q_i
  \end{bmatrix}^{\top},
  \qquad
  \boldsymbol{\beta}
  =
  \begin{bmatrix}
    \mu & b & I
  \end{bmatrix}^{\top}.
\end{equation}
Stacking the retained regressors into $Z$ gives
\begin{equation}
  \label{eq:nnls}
  \hat{\boldsymbol{\beta}}
  =
  \underset{\boldsymbol{\beta}\geq0}
  {\operatorname{argmin}}
  \left\|
    \boldsymbol{\tau}
    -
    Z\boldsymbol{\beta}
  \right\|_2^2 .
\end{equation}
The elementwise nonnegativity constraint enforces nonnegative effective
inertia, Coulomb friction, and viscous damping. These quantities correspond to
properties commonly supported by rigid-body simulators, although their exact
interpretation and implementation remain backend-dependent.
We next fit the structured nonlinear model to the residual
\begin{equation}
  r_i
  =
  \tau_i-z_i^\top\hat{\boldsymbol{\beta}},
\end{equation}
using $  \hat r_i
  =
  \tau_{\mathrm{mech}}(q_i,\dot q_i),
  \qquad
  c_\phi\geq0.$
The configuration-dependent head $g_\theta$ is sign-unconstrained and takes
only $q$ as input. The damping coefficient is produced by softplus outputs,
with separate heads for positive and negative motion, allowing directional
asymmetry while preserving nonnegative damping.
The resulting model is partially linear, but its decomposition is not
generally unique. In particular, a constant component of $c_\phi$ is
indistinguishable from the linear damping coefficient $b$, and
configuration-dependent gravity is indistinguishable from a mechanical load
represented by $g_\theta$. Correlations between $q$, $\dot q$, and
$\ddot q$ can introduce further coupling between the fitted terms. We
therefore interpret $\hat{\boldsymbol{\beta}}$ as effective
simulator-compatible parameters and evaluate the predictive fidelity of the
complete effort model. The structured residual and probing protocol regularize
the decomposition, but do not make every component uniquely identifiable.
We fit the model sequentially: nonnegative least squares followed by
residual-network training. Training minimizes residual MSE using full-batch
AdamW (learning rate and weight decay $10^{-3}$), with a random
75/25 training/validation split, at most 1200 epochs, and early stopping
after 120 epochs without validation improvement; we retain the
best-validation weights for the mechanism term.

\subsection{Twin Assembly}

\begin{figure}[tb]
  \centering
  \includegraphics[width=0.95\columnwidth,
  trim={1cm 0.5cm 1cm 0.5cm},
  clip
]{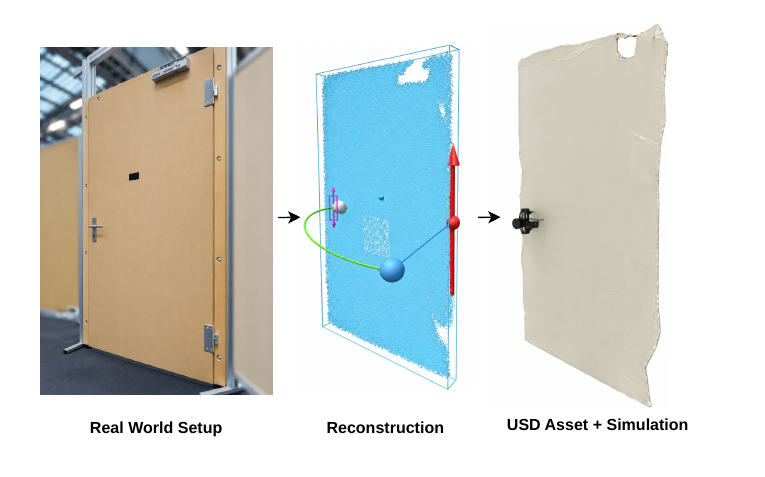}
      \vspace{-5pt}
  \caption{\textbf{Construction of a physics-informed twin.}
    (a) Reconstructed panel instance, associated handle, and registered joint
    axis. (b) Exported articulated twin. The physical model is transformed
    into the asset frame, reparameterized to match its joint-coordinate
    convention, and attached to the corresponding articulated instance.}
        \vspace{-15pt}
  \label{fig:twin-3d-structure}
\end{figure}

The identified physical model is expressed in the probing frame $W$ and joint
coordinate defined above. To attach it to a target geometric representation
with frame $A$, we first register the probing frame using
${}^{A}T_W$. The joint twist transforms according to ${}^{A}\boldsymbol{\xi} = \operatorname{Ad}_{{}^{A}T_W}{}^{W}\boldsymbol{\xi}.$
Rigid frame changes preserve the scalar joint coordinate and generalized
effort. If the target asset instead uses a different coordinate convention %
\begin{equation}
  q_A=sq+q_{\mathrm{off}},
  \qquad s\in\{-1,1\},
\end{equation}
we transform the valid range and reparameterize the mechanism model.
For the complete mechanism effort
$\tau_{\mathrm{mech}}(q,\dot q)$, the target convention uses $\tau_{\mathrm{mech}}^{A}(q_A,\dot q_A) = s\,\tau_{\mathrm{mech}}\bigl(s(q_A-q_{\mathrm{off}}),\, s\dot q_A\bigr)$ .
The nonnegative magnitudes $I$, $\mu$, and $b$ are unchanged under the sign
reversal, while their associated signed efforts follow the transformed joint
coordinate.
We then associate the registered model with the corresponding articulated
instance, as illustrated in \cref{fig:twin-3d-structure}. The 
parametric dynamics are mapped to the corresponding joint or actuator properties
provided by the target backend. When the backend supports runtime
joint-effort evaluation, the nonlinear model is evaluated directly. For
backends that cannot execute the learned network, we approximate it with a
backend-supported spring-damper representation during export. This conversion
is distinct from physical identification and may introduce an additional
approximation.
In our experiments, the geometric representation is reconstructed from the
same recording, so it already shares the probing frame. We associate the
manipulated handle with the tracked TCP and use the reconstructed functional
relation to identify the corresponding panel. Other geometric backends or
existing assets require only a registration and an instance-association
procedure. The resulting twin can be consumed either as a feedforward
dynamics model for robot control or as an articulated simulation asset.

\msection{Experiments}\label{sec:experiments}

Evaluation proceeds along two axes. First, we ask whether the estimated parameters are accurate and whether that accuracy justifies the model's complexity: \cref{sec:est-fidelity} measures fidelity against ground truth inertia measurements and real-world behavior, and \cref{sec:ablations} investigates the contribution of each model component. Second, we investigate the utility of ForceTwin: \cref{sec:manip} deploys the twin as a feedforward dynamics model on two physical robots, and \cref{sec:rl} exports it as a simulation asset for policy learning. All experiments follow the capture protocol of \cref{sec:setup}.

\subsection{Capture Protocol and Implementation}\label{sec:setup}

\begin{figure}[tb]
  \centering
  \includegraphics[width=\columnwidth]{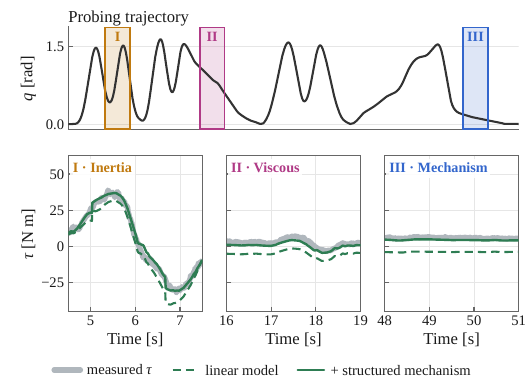}
      \vspace{-20pt}
  \caption{\textbf{Dynamics model fit for the wood door.} (a) Measured
    generalized effort (gray) against the parametric model alone (dashed) and with
    the structured mechanism added (green); (b) the probing trajectory. The
    parametric model tracks the shape of the effort but is offset from it by a
    configuration-dependent amount, which the learned mechanism recovers.
    Shaded bands mark the probing regimes for inertia (yellow), viscous damping (purple), and the mechanism 
     (blue).}
        \vspace{-15pt}
  \label{fig:dynamics-model-fit-probing-rsl-door}
\end{figure}

We capture all interactions with the handheld Hoi!
gripper~\cite{Engelbracht_2026_CVPR}, which provides posed RGB-D
observations, the tool trajectory from Project Aria
MPS~\cite{engel2023projectarianewtool}, and the contact wrench from a
wrist-mounted force-torque sensor. %
Once a part is grasped, we open and close it over a range of speeds and accelerations chosen to excite
the individual regressor axes: slow, low-acceleration motion for Coulomb
friction, a wide speed spectrum for viscous damping, high accelerations for
inertia, and passive following for the mechanism.
\Cref{fig:dynamics-model-fit-probing-rsl-door}(b) shows an example probing
sequence. In our implementation,
FunFact~\cite{fu2026funfactbuildingprobabilisticfunctional} reconstructs
panels, handles and functional relations from the same recording and
therefore already shares the probing frame; we identify the manipulated
handle by proximity to the tracked TCP, use FunFact's handle-to-panel
relation to select the associated door or drawer panel, attach the recovered
model to that pair, process the instance meshes, and export the articulated
asset to USD. Because $\mathcal{M}_{\mathrm{phys}}$ is independent of the
geometric representation, it can equally be registered onto an existing
asset: the real-to-sim free-swing experiments of \cref{sec:est-fidelity} use
the exported asset, while the policy-learning experiments of \cref{sec:rl}
attach the identified dynamics and mechanism to the door asset of the
training pipeline, leaving its geometry unchanged. In both cases the
identified $\hat{\boldsymbol{\beta}}$ is written directly to the joint's
inertia, friction and damping fields, so the parametric half of the twin is
native to the simulator, while the mechanism term is evaluated per step as a
state-dependent joint effort. An independently obtained representation has to
be registered to the probing frame first.

\subsection{Estimation Fidelity}\label{sec:est-fidelity}

We evaluate estimation accuracy at three levels: kinematic parameters, compared against annotated ground-truth axes on the real-world test rig; inertial parameters, whose ground truth can be independently measured (unlike friction parameters, for which no such reference exists); and finally aggregate system fidelity, where kinematic and dynamic terms are evaluated jointly.

\PAR{Kinematics Estimation.}
We evaluate articulation types and axes on our real-world test bench against ground truth annotated in a Leica RTC360 laser scan~\cite{Engelbracht_2026_CVPR,werby2025articulatedobjectestimationwild}.
After registering estimates to the scan, we measure axis-angle errors for both joint types and axis-position errors for revolute joints~\cite{werby2025articulatedobjectestimationwild,büchner2026articulated3dscenegraphs}.
We compare against MoMa-SG~\cite{büchner2026articulated3dscenegraphs}, which tracks object points during non-instrumented human interaction and is run on our interactions, and OPDMulti~\cite{sun2023opdmultiopenabledetectionmultiple}, which estimates articulation from visual input without interaction. Interaction-based methods outperform the visual baseline (\cref{tab:axis-main-sd}).
OPDMulti estimates prismatic directions reasonably well ($4.8^\circ$), but yields large revolute-axis errors ($34.9^\circ$) and only $54.1\%$ revolute type accuracy; both interaction-based methods classify all joints correctly.
Compared with MoMa-SG, our method slightly improves revolute-axis orientation ($1.5^\circ$ vs.\ $1.7^\circ$) and halves position error ($23$ vs.\ $46$\,mm).
We attribute this improvement to directly measured tool poses, avoiding reliance on surface-point tracking susceptible to low texture and motion blur.

\begin{table}[tb]
  \centering
  \caption{\label{tab:axis-main-sd}\tabtitle{Articulation Axis
    Estimation.} Evaluated on the real-world test bench against 12 axes
    annotated in a Leica RTC360 laser scan. Metrics follow
    ArtiPoint~\cite{werby2025articulatedobjectestimationwild}:
    $\theta_\mathrm{err}$ is the angle between axis lines, folded to
    $[0,90]^\circ$, and $d_{L_2}$ the distance between revolute axis lines
    along their common perpendicular. Errors accumulate only over
    predictions with the correct joint type. Values are means with the
    sample standard deviation in smaller type; $^\dagger$no deviation
    available. \textbf{Bold} indicates best performance,
    \underline{underline} second best; tied entries are all marked best.}
  \footnotesize
  \setlength{\tabcolsep}{3pt}
  \rowcolors{3}{gray!10}{white}
  \resizebox{\columnwidth}{!}{%
  \begin{tabular}{l r r r r r}
    \toprule
    & \textbf{Prismatic} & \multicolumn{2}{c}{\textbf{Revolute}} & \multicolumn{2}{c}{\textbf{Type acc.} [\%] $\uparrow$} \\
    \cmidrule(lr){2-2} \cmidrule(lr){3-4} \cmidrule(lr){5-6}
    \textbf{Method} & $\theta_\mathrm{err}$ [$^\circ$] $\downarrow$ & $\theta_\mathrm{err}$ [$^\circ$] $\downarrow$ & $d_{L_2}$ [mm] $\downarrow$ & \textbf{Pris.} & \textbf{Rev.} \\
    \midrule
    \OurMethod{} (Ours) & \sd{\textbf{1.4}}{0.2} & \sd{\textbf{1.5}}{0.8} & \sd{\textbf{23}}{26} & \textbf{100.0} & \textbf{100.0} \\
    MoMa-SG~\cite{büchner2026articulated3dscenegraphs} & \sd{\underline{1.6}}{0.8} & \sd{\underline{1.7}}{0.0} & \sd{\underline{46}}{20} & \textbf{100.0} & \textbf{100.0} \\
    OPDMulti~\cite{sun2023opdmultiopenabledetectionmultiple} & 4.8$^\dagger$\phantom{\,{\tiny$\pm$0.0}} & \sd{34.9}{29.1} & \sd{143}{117} & \textbf{100.0} & 54.1 \\
    \bottomrule
  \end{tabular}}
      \vspace{-10pt}
\end{table}

\PAR{Dynamics and Mechanism Estimation.}
We next assess the fidelity of our estimated dynamics parameters against a
prior-based paradigm such as ~\cite{iliash2026artiverse}. We focus on the inertial
parameters, since ground truth is readily obtainable for them. For prismatic
joints this is the mass, measured with a scale. For revolute panels, the
hinge inertia follows from mass and geometry as $I = \tfrac{1}{3} m \ell^2$ under
the assumption of constant density, while compact parts such as locks and
handles are treated as point masses with $I = m \ell^2$. Here $\ell$ is
the distance from the hinge axis: the panel's extent from the hinge in the
first case, and the radial offset of the part in the second.
No ground truth is available for the parametric friction parameters or for the
nonlinear MLP. To still evaluate these terms, we
assess the aggregate fit at the system level: the real-to-sim free-swing
experiment below compares the closing behavior of the real
object with that of its simulated twin.
\Cref{fig:dynamics-model-fit-probing-rsl-door} shows a qualitative example of
the fit. The parametric model tracks the measured effort qualitatively, but with a non-constant offset that reveals the mechanism it cannot represent.
\begin{figure}[tb]
\vspace{-4pt}
  \centering
  \includegraphics[width=0.9\columnwidth]{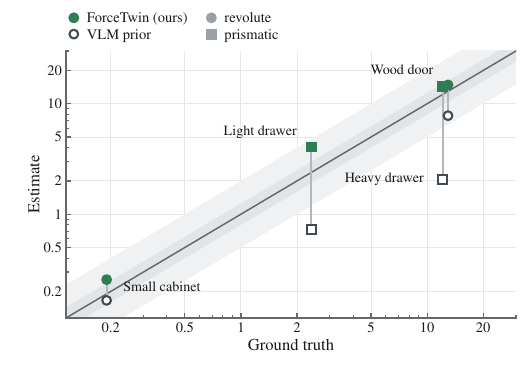}
      \vspace{-2pt}
  \caption{\textbf{Inertial-parameter fidelity on the four objects for which
    ground truth is available:} estimate against ground truth on logarithmic
    axes, so an estimate on the diagonal is exact and vertical distance from
    it is the error. Marker shape distinguishes revolute (circle) from
    prismatic (square) joints, whose parameters carry the two units given on
    the axes. Nested shaded bands mark agreement
    to within $\pm25\%$ (inner) and to within a factor of two (outer). %
    The macro mean of the absolute
    relative errors is $29.5\%$ for \OurMethod{} and $53.0\%$ for the VLM
    prior.}
        \vspace{-15pt}
  \label{fig:inertia-fidelity}
\end{figure}
Aggregated over all objects (\cref{fig:inertia-fidelity}), the interaction-based estimate is more accurate than the prior-based one ($29.5\%$ vs.\ $53.0\%$ macro mean
absolute relative error),
and it is the better estimate on three of the four objects for which ground
truth is available. Our accuracy improves systematically with the magnitude of
the inertial parameter: the two heaviest parts are estimated to within
$10.4\%$ and $13.7\%$, whereas the lightest ones reach $63.8\%$ and
$30.2\%$. We attribute this to the signal-to-noise ratio of the measurement rather than to
the estimator itself, as light parts excite only small interaction forces, so
that the force-torque noise band occupies a larger fraction of the measured
signal. Consistently, the sole case in which the prior-based baseline is more
accurate is the smallest object in the set.
The prior-based baseline underestimates the inertial parameter for
every object, despite being provided with scale information and prompted
to reason about material. The largest error occurs for the heavy drawer
($83.5\%$), whose effective mass is dominated by its contents, hidden state that
cannot be inferred from appearance.
Our estimator, in turn, overestimates on all four objects. The bias is
consistent in sign, which suggests that residual unmodeled effort is
partially absorbed into the inertial term during the parametric fit. This
reflects a structural property of the decomposition rather than insufficient
optimization: both the parametric terms and the mechanism response depend on the
articulation velocity, so the split between them is only partially
identifiable from torque observations, and the fit objective is predictive
accuracy rather than parameter recovery. The downstream uses of the twin
consume the full torque model, whose aggregate fidelity is assessed at
system level by the free-swing experiment below.

\PAR{System-Level Fidelity: Real-to-Sim Free Swing.}
The preceding experiment evaluates individual parameters, but only the inertial
ones admit ground truth. We therefore assess the twin at the system level,
where inertia, friction and mechanism act jointly. We release the metal door and the wood door from $45^\circ$ and $90^\circ$ and compare the measured closing time
against simulations parameterized either by our interaction-derived estimates
or by the VLM-prior baseline.
\begin{table*}[!t]
\centering
\begin{minipage}[t]{0.68\textwidth}
\vspace{0pt}
\centering
\caption{
\tabtitle{Real-Robot Articulation.}
Spot and Franka FR3 track robot-specific opening goals within 6\,s using a fixed controller.
\OurMethod{} uses interaction-identified dynamics; VLM prior uses visual and language estimates~\cite{iliash2026artiverse,li2026physgraphphysicsaware3dscene}; Kinematics only omits dynamics compensation~\cite{büchner2026articulated3dscenegraphs,Huang_2026}.
\emph{Complete}: maximum goal completion (capped at 100\%).
\emph{Track}: goal-normalized tracking RMSE (\%).
Values are means with sample SD; the overall mean equally weights nine object--robot pairs.
\emph{Mechanism} rates the contribution of built-in mechanisms to articulation force.
Objects are ordered by \OurMethod{} tracking error within each robot panel.
\textbf{Bold}/\underline{underline}: best/second best per object and metric, with ties marked best.}
\vspace{-14pt}
\label{tab:real-robot-6s}
\setlength{\tabcolsep}{3pt}
\resizebox{.85\linewidth}{!}{%
\begin{tabular}{@{}c@{\hspace{4pt}}llrr|rr|rr@{}}
\toprule
& & & \multicolumn{2}{c}{\textbf{\OurMethod{} (Ours)}}
& \multicolumn{2}{c}{\textbf{VLM prior}}
& \multicolumn{2}{c}{\textbf{Kinematics only}} \\
\cmidrule(lr){4-5}\cmidrule(lr){6-7}\cmidrule(lr){8-9}
& \textbf{Object} & \textbf{Mechanism}
& \textbf{Complete} $\uparrow$ & \textbf{Track} $\downarrow$
& \textbf{Complete} $\uparrow$ & \textbf{Track} $\downarrow$
& \textbf{Complete} $\uparrow$ & \textbf{Track} $\downarrow$ \\
\midrule
\multirow{5}{*}{\rotatebox[origin=c]{90}{\textbf{Spot}}}
& Drawer & Weak
& \sd{88.0}{1.8} & \sd{\textbf{9.2}}{2.9}
& \sd{\textbf{95.2}}{4.0} & \sd{\underline{34.1}}{3.1}
& \sd{\underline{90.6}}{3.7} & \sd{38.0}{2.6} \\
& Sliding door & None
& \sd{\textbf{100.0}}{0.0} & \sd{\textbf{10.5}}{1.9}
& \sd{\textbf{100.0}}{0.0} & \sd{\underline{15.4}}{1.9}
& \sd{\textbf{100.0}}{0.0} & \sd{18.5}{2.6} \\
& Wood door & Medium
& \sd{\textbf{100.0}}{0.0} & \sd{\textbf{14.0}}{1.5}
& \sd{\underline{91.2}}{0.9} & \sd{\underline{24.3}}{0.0}
& \sd{88.3}{4.7} & \sd{27.3}{1.6} \\
& Oven & Strong
& \sd{\textbf{100.0}}{0.0} & \sd{\textbf{33.3}}{3.1}
& \sd{\underline{4.0}}{0.7} & \sd{\underline{91.1}}{0.4}
& \sd{2.2}{0.1} & \sd{91.9}{0.2} \\
& Metal door & Strong
& \multicolumn{6}{c}{N/A (object forces exceed Spot's actuation limits)} \\
\midrule
\multirow{5}{*}{\rotatebox[origin=c]{90}{\textbf{Franka}}}
& Sliding door & None
& \sd{93.7}{0.7} & \sd{\textbf{10.5}}{0.4}
& \sd{\underline{98.7}}{1.0} & \sd{\underline{17.2}}{0.9}
& \sd{\textbf{99.3}}{1.0} & \sd{26.9}{12.2} \\
& Drawer & Weak
& \sd{\underline{85.6}}{9.6} & \sd{\textbf{17.6}}{3.2}
& \sd{\textbf{93.1}}{3.2} & \sd{37.0}{0.4}
& \sd{85.1}{11.2} & \sd{\underline{22.4}}{4.2} \\
& Metal door & Strong
& \sd{\textbf{81.2}}{3.5} & \sd{\textbf{24.4}}{5.5}
& \sd{\underline{1.6}}{0.0} & \sd{\underline{91.0}}{0.0}
& \sd{1.5}{0.0} & \sd{\underline{91.0}}{0.0} \\
& Wood door & Medium
& \sd{\textbf{69.8}}{19.4} & \sd{\textbf{30.3}}{11.0}
& \sd{\underline{51.7}}{4.0} & \sd{\underline{47.7}}{2.8}
& \sd{44.0}{21.3} & \sd{56.0}{16.6} \\
& Oven & Strong
& \sd{\textbf{67.6}}{0.0} & \sd{\textbf{49.7}}{0.0}
& \sd{\underline{1.8}}{1.0} & \sd{\underline{90.8}}{0.9}
& \sd{0.9}{0.1} & \sd{91.6}{0.1} \\
\midrule
\rowcolor{gray!15}
\multicolumn{3}{l}{\textbf{Overall mean}}
& \textbf{87.3} & \textbf{22.2}
& \underline{59.7} & \underline{49.8}
& 56.9 & 51.5 \\
\bottomrule
\end{tabular}%
}
\end{minipage}\hfill%
\begin{minipage}[t]{0.3\textwidth}
\vspace{0pt}
\centering
\caption{
\tabtitle{Real-to-Sim Free-Swing Fidelity.}
Closing times of the physical door and of twins parameterized by our
interaction-derived estimates and by the VLM prior, released from two
initial opening angles $q_0$. Times are in seconds and the absolute
relative error (ARE) is a percentage. The final row is the equal-weight
macro mean. Only two estimates are compared, so \textbf{bold} marks
the lower ARE and no second rank is reported.}
\label{tab:real-to-sim-free-swing}
\footnotesize
\setlength{\tabcolsep}{2pt}
\rowcolors{3}{gray!10}{white}
\resizebox{\linewidth}{!}{%
\begin{tabular}{@{}lcccccc@{}}
\toprule
& & & \multicolumn{2}{c}{\textbf{ForceTwin}}
& \multicolumn{2}{c}{\textbf{VLM prior}} \\
\cmidrule(lr){4-5}\cmidrule(lr){6-7}
\textbf{Object} & $q_0$ [$^\circ$] & \textbf{Real} [s]
& $t$ [s] & \textbf{ARE} $\downarrow$
& $t$ [s] & \textbf{ARE} $\downarrow$ \\
\midrule
Metal door & 45 & 5.6 & 5.9 & \textbf{5.4} & 2.1 & 62.5 \\
Metal door & 90 & 8.4 & 8.8 & \textbf{4.8} & 2.5 & 70.2 \\
Wood door & 45 & 3.7 & 5.0 & \textbf{35.1} & 2.0 & 45.9 \\
Wood door & 90 & 5.0 & 6.0 & \textbf{20.0} & 2.4 & 52.0 \\
\midrule
Macro mean & -- & -- & -- & \textbf{16.3} & -- & 57.7 \\
\bottomrule
\end{tabular}%
}
\end{minipage}
\vspace{-10pt}
\end{table*}
Our estimates yield the lower error in all four conditions
(\cref{tab:real-to-sim-free-swing}), reducing the macro
mean from $57.7\%$ to $16.3\%$. The VLM-prior baseline closes every door in roughly
$2$\,s regardless of door or release angle, indicating that it does not match the
damped dynamics closely. Note that this measure is aggregate and not injective: a large mismatch proves the dynamics are wrong, but a close match does not uniquely identify the individual parameters, since compensating errors could in principle yield the same closing time. It therefore complements the parameter-wise evaluation above.

\subsection{Model-Class Ablations}\label{sec:ablations}
We evaluate the benefit of added model complexity. Adding viscous damping to the Coulomb-inertia model reduces in-sample effort RMSE by up to $12\%$. Adding the structured mechanism reduces it by a further $67-73\%$ on mechanism-dominated parts such as the doors and the oven, against $12\%$ on
the sliding door, which has no mechanism. We fit the two stages sequentially, the parametric coefficients first and the mechanism on their residual. Alternating between them changes the fit RMSE by less than $5\%$ of the parametric baseline while redistributing effort between overlapping terms, so we do not iterate.

\subsection{Does Dynamics Fidelity Change Manipulation?}\label{sec:manip}

\begin{figure}[tb]
\vspace{-4pt}
  \centering
  \includegraphics[width=0.85\columnwidth]{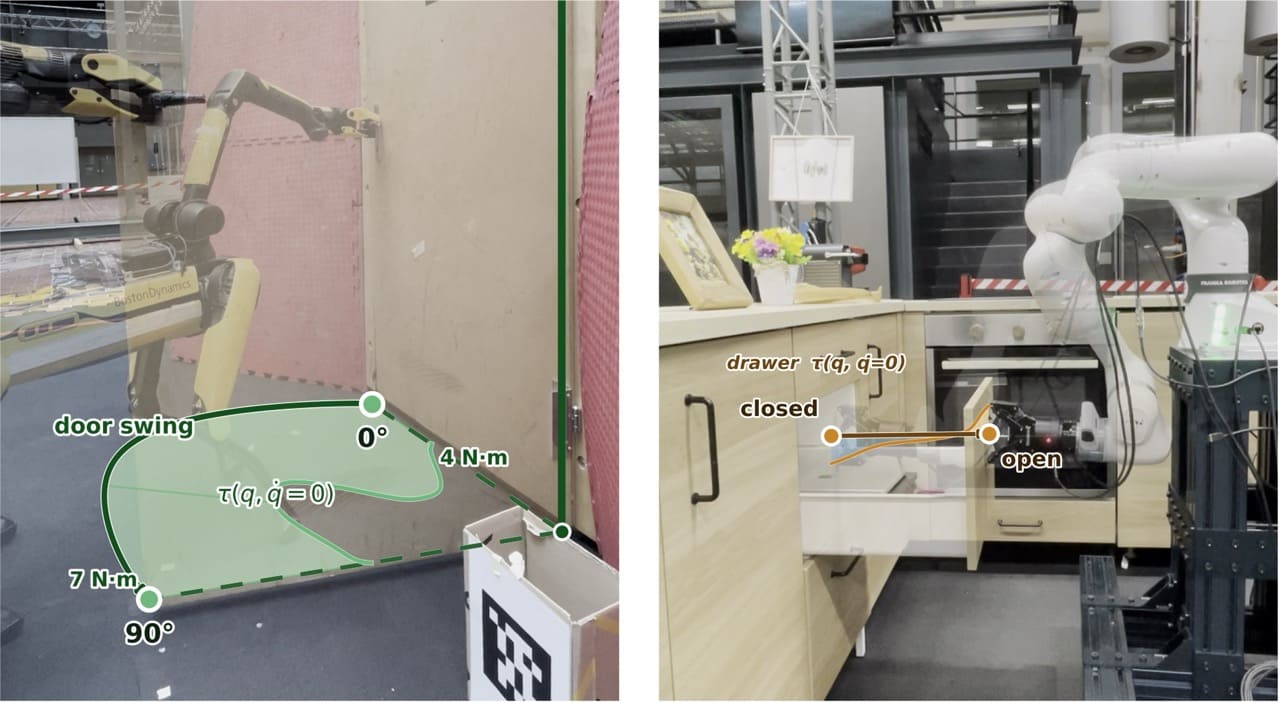}
\caption{\textbf{Real-world experiments of zero-shot model-based articulation }(\cref{tab:real-robot-6s}) across two embodiments: Spot and Franka FR3. We visualize the initial and final articulation states as blended overlays for both embodiments. The recovered quasi-static mechanism torque for each object is plotted over the opening trajectory, highlighting the variation in mechanism behavior across objects.} %
    \vspace{-18pt}
  \label{fig:experiment-door-rsl}
\end{figure}

This experiment evaluates whether the fidelity of a twin's dynamics influences manipulation performance: We use ForceTwin's object dynamics as a feedforward term of a Cartesian impedance controller,\vspace{-4pt}
\begin{equation}
\begin{aligned}
  \mathbf{f}_{\mathrm{cmd}}
  &= K_x(x_d-x)
   + D_x(\dot{x}_d-\dot{x})
   + \mathbf{f}_{\mathrm{ff}}, \\
  \mathbf{f}_{\mathrm{ff}}
  &= \left(J_{p}(q)^{\top}\right)^{\dagger}
  \left[
    \hat I\ddot q_d
    + \hat\mu\,\operatorname{sgn}(\dot q_d)
    + \hat b\dot q_d
    + \hat\tau_{\mathrm{mech}}(q_d,\dot q_d)
  \right].
\end{aligned}
\end{equation}
\vspace{-4pt}
where the bracket sums the generalized effort of the identified inertia, Coulomb and viscous damping, and mechanism response along the reference articulation trajectory $q_d(t)$, and the transposed pseudoinverse of the object Jacobian's linear block $J_{p}$ maps them to an end-effector force added to the impedance feedback. The twin enters only through $\mathbf{f}_{\mathrm{ff}}$, so we vary its source across the two prevalent twin-construction paradigms and ours, keeping robot, controller, and task fixed: kinematics-only twins~\cite{sun2023opdmultiopenabledetectionmultiple,Huang_2026,büchner2026articulated3dscenegraphs,delitzas2026funrecreconstructingfunctional3d} provide $J_{p}$ but no dynamics ($\mathbf{f}_{\mathrm{ff}}=\mathbf{0}$), VLM-prior twins~\cite{iliash2026artiverse,li2026physgraphphysicsaware3dscene} supply prior-based parameter estimates to the same feedforward, and \OurMethod{} supplies the parameters identified from instrumented interaction. In each trial, the robot starts with the handle grasped and tracks $q_d(t)$ from the closed configuration to an embodiment-specific goal within 6\,s (\cref{fig:experiment-door-rsl}); We report goal completion and goal-normalized tracking RMSE.
\Cref{tab:real-robot-6s} reports goal completion and tracking error on five
objects across Spot and Franka FR3 under a common 6\,s goal. \OurMethod{}
yields the lowest tracking error on every object and the highest macro mean
completion ($87.3\%$, against $59.7\%$ for the VLM prior and $56.9\%$ for
kinematics only), and the margin scales with the dynamic load. On the
drawers
and sliding doors all three models complete the goal, and the identified
model
improves tracking error by factors of $1.3$ to $4$. On the strong-mechanism
objects, the oven on both robots and the metal door on the Franka, both
baselines stall below $5\%$ completion because their feedforward supplies
only
a fraction of the required force, while the identified model reaches
$67$--$100\%$. The wood door lies between these regimes on both
embodiments,
where the baselines lose $9$--$26$ percentage points and roughly double the
tracking error.

\vspace{-4pt}
\subsection{The Twin as a Training Asset}\label{sec:rl}

\begin{figure}[t]
\vspace{-14pt}
    \centering
    \includegraphics[width=0.95\linewidth  ,
  trim={5mm 0.0cm 0cm 0.0cm},
  clip]{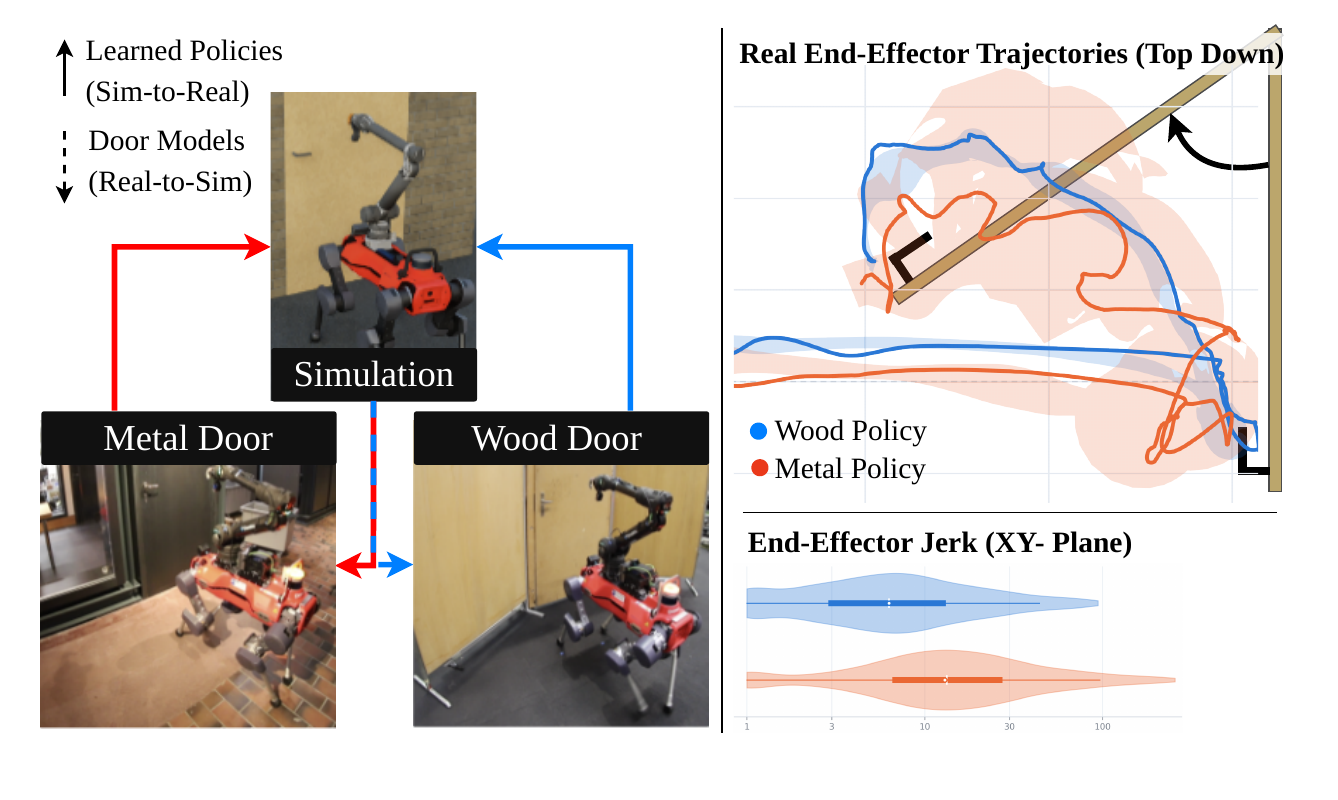}
\vspace{-18pt}
\caption{\textbf{Policy learning with identified door models.}
\textbf{Left:} We use the identified wood- and metal-door models (real-to-sim) to train whole-body policies, which we then deploy on ANYmal (sim-to-real).
\textbf{Right:} Both policies on the real wood door; shaded regions show the average over five trials. The wood-door policy moves the end-effector smoothly, while the metal-door policy stops and starts, pulling hard and then pausing, which gives it higher jerk (bottom) due to the worse sim-to-real calibration.}
\vspace{-15pt}
    \label{fig:rsl-r2s2}
\end{figure}

Beyond the feedforward evaluations above, we show that the identified models
can be used directly for reinforcement learning. We register the identified wood- and metal-door dynamics, including the recovered nonlinear closing mechanism, onto the door asset in Isaac Lab~\cite{nvidia2025isaaclabgpuacceleratedsimulation}. Using the motion-imitation pipeline from~\cite{sleiman2024guidedreinforcementlearningrobust},  we train whole-body door-traversal policies. That pipeline randomizes the door parameters during training; here we replace that randomization with the identified dynamics, leaving the policy architecture and training procedure unchanged.
We deploy both policies on an ANYmal quadruped with an arm and evaluate them on the two real doors. The wood-door policy succeeds on the wood door in all five trials. The metal-door policy also succeeds on this door, but exhibits a stop-and-go motion: it advances the door and then pauses while the imitation phase catches up (\cref{fig:rsl-r2s2}). On the metal door, the metal-door policy succeeds, whereas the wood-door policy fails its initial opening attempt.
These experiments demonstrate that the identified models integrate into an existing policy-learning pipeline and support real-world deployment on contact-rich whole-body tasks. Whether training with identified dynamics consistently improves performance over domain randomization requires a controlled comparison, which we leave to future work.

\vspace{-5pt}
\msection{Limitations}

Our decomposition of the effort model is not unique: a constant mechanism
damping is absorbed by $b$, and any configuration-dependent load by
$g_\theta$, including gravity when the joint axis is not vertical. 
The model is memoryless in $(q,\dot q)$, so hysteresis, backlash, stiction at rest, and latch states lie outside its class, and both $g_\theta$ and the joint range
hold only where the object was probed. Further, identification is per instance and needs physical interaction: each part is probed individually, and the probing has to excite the individual regressor axes, so passive demonstration recordings do not suffice. Priors scale across a scene where probing does not, so the two are complementary, and transferring identified dynamics to parts that were not probed is left to future work. Identification is also one-shot: a loaded drawer or a readjusted closer changes the object, and refining a twin online from the robot's own
interaction forces remains open.

\vspace{-5pt}
\msection{Conclusion}

\OurMethod{} identifies physics-informed digital twins of articulated objects from instrumented human interaction. A person probes the object with a handheld force-sensing gripper, and from the tool trajectory and contact wrenches we recover the articulation, effective inertia, Coulomb and viscous terms under nonnegativity constraints, and a residual for the nonlinear mechanism. The identified parameters nearly halve the inertial error of a prior-based estimate and reproduce measured door-closing times. A single twin then serves two uses: as the feedforward model of an impedance controller, it operates objects on which kinematics-only and prior-based twins stall, and exported to simulation it is used to train whole-body door-traversal policies transferring to the real objects. 

\vspace{-5pt}
\section*{\small ACKNOWLEDGEMENTS}
 
{\small
This work was supported by SNSF Advanced Grant 216260, the Lamarr Institute,
and Google. The authors used generative AI to assist with
drafting and editing the manuscript, cleaning up figures and coding; all
content was reviewed by the authors, who take full responsibility for it.
\par}

\ifprintbibliography
\vspace{-12pt}
  \bibliographystyle{IEEEtran}
  \bibliography{references}
\fi

\end{document}